%% file: root.tex
\documentclass[letterpaper, 10 pt, conference]{inc/ieeeconf}  

\IEEEoverridecommandlockouts                              

\usepackage{subcaption}
\usepackage{textcomp}
\usepackage{xcolor} 
\usepackage{pifont}
\usepackage{rotating}
\usepackage{graphicx} 
\usepackage{booktabs} 
\usepackage{multirow} 
\usepackage{tabularx} 
\usepackage{adjustbox}
\usepackage[section]{placeins}
\usepackage{caption}
\usepackage{graphicx}
\usepackage{xurl}
\usepackage[bookmarks=true,breaklinks=true,letterpaper=true,colorlinks,linkcolor=blue,citecolor=blue,urlcolor=cyan]{hyperref}
\usepackage[separator = {;}]{credits}
\usepackage{siunitx}

\title{\LARGE \bf
    RobotPerf: An Industry-Strength Robotics Benchmark Methodology for ROS~2 across Hardware Platforms
}

\title{\LARGE \bf
RobotPerf: A Comprehensive Benchmarking Methodology for Evaluating ROS 2 Performance on Heterogeneous Hardware Platforms}

\title{\LARGE {\bf
RobotPerf}: An Open-Source, Vendor-Agnostic, Benchmarking Suite\\for Evaluating Robotics Computing System Performance}

\author{Víctor Mayoral-Vilches$^{1,2}$, Jason Jabbour$^{3}$, Yu-Shun Hsiao$^{3}$, Zishen Wan$^{4}$, Martiño Crespo-Álvarez$^{1}$, \\ Matthew Stewart$^{3}$, 
    Juan Manuel Reina-Muñoz$^{1}$, Prateek Nagras$^{1}$, Gaurav Vikhe$^{1}$,\\ Mohammad Bakhshalipour$^{5}$, Martin Pinzger$^{2}$, Stefan Rass$^{6,2}$, Smruti Panigrahi$^{7}$, Giulio Corradi$^{8}$,\\ Niladri Roy$^{9}$, Phillip B. Gibbons$^{5}$, Sabrina M. Neuman$^{10}$, Brian Plancher$^{11}$, Vijay Janapa Reddi$^{3}$
    \vspace{-1em}
\thanks{$^{1}$\href{https://accelerationrobotics.com/}{Acceleration Robotics}, Spain. $^{2}$\href{https://aau.at/}{Alpen-Adria-Universität Klagenfurt}, Austria. $^{3}$\href{https://harvard.edu}{Harvard University}, USA. $^{4}$\href{https://www.gatech.edu/}{Georgia Institute of Technology}, USA. $^{5}$\href{https://www.cmu.edu/}{Carnegie Mellon University}, USA. $^{6}$\href{https://www.jku.at/}{Johannes Kepler University Linz}, Austria. $^{7}$\href{https://ford.com}{Ford Motor Company}, USA. $^{8}$\href{https://www.amd.com/}{AMD}, USA. 
$^{9}$\href{https://intel.com/}{Intel}, USA.
$^{10}$\href{https://www.bu.edu/cs/}{Boston University}, USA. $^{11}$\href{https://cs.barnard.edu}{Barnard College, Columbia University}, USA.}
}

\newcommand{\cmark}{\textcolor{green!80!black}{\ding{51}}}

\usepackage{flushend}

\newcommand{\xmark}{\textcolor{red}{\ding{55}}}

\definecolor{robotperfblue}{HTML}{4d4cf5}

\begin{document}

\maketitle
\thispagestyle{empty}
\pagestyle{empty}

\begin{abstract}
\input{tex/0_abstract}
\end{abstract}

\vspace{-.5em}
\section{Introduction}
\label{sec:intro}
\input{tex/1_intro}

\section{Background \& Related Work}
\label{sec:background_motivation}
\input{tex/2_background}

\input{tex/3_related}

\input{tex/4_design}
\label{sec:design}


\section{Evaluation}
\label{sec:eval}
\input{tex/5_eval}






\bibliographystyle{inc/IEEEtran}
\bibliography{references}

\end{document}

%% file: tex/0_abstract.tex
%
We introduce RobotPerf, a vendor-agnostic benchmarking suite designed to evaluate robotics computing performance across a diverse range of hardware platforms using ROS 2 as its common baseline. The suite encompasses ROS 2 packages covering the full robotics pipeline and integrates two distinct benchmarking approaches: black-box testing, which measures performance by eliminating upper layers and replacing them with a test application, and grey-box testing, an application-specific measure that observes internal system states with minimal interference. Our benchmarking framework provides ready-to-use tools and is easily adaptable for the assessment of custom ROS 2 computational graphs. Drawing from the knowledge of leading robot architects and system architecture experts, RobotPerf establishes a standardized approach to robotics benchmarking. As an open-source initiative, RobotPerf remains committed to evolving with community input to advance the future of hardware-accelerated robotics. 

%% file: tex/1_intro.tex

In order for robotic systems to operate safely and effective in dynamic real-world environments, their computations must run at real-time rates while meeting power constraints.
Towards this end, accelerating robotic kernels on heterogeneous hardware, such as GPUs and FPGAs, is emerging as a crucial tool for enabling such performance~\cite{neuman2021robomorphic,liu2021archytas,makoviychuk2021isaac,plancher2022grid,mayoral2022robotcore,wan2022robotic,liu2021robotic}. This is particularly important given the impending end of Moore's Law and the end of Dennard Scaling, which limits single CPU performance~\cite{Esmaeilzadeh11,Venkatesh10}.

While hardware-accelerated kernels offer immense potential, they necessitate a reliable and standardized infrastructure to be effectively integrated into robotic systems. As the industry leans more into adopting such standard software infrastructure, the Robot Operating System (ROS)~\cite{quigley2009ros} has emerged as a favored choice. Serving as an industry-grade middleware, it aids in building robust computational robotics graphs, reinforcing the idea that robotics is more than just individual algorithms. The growing dependency on ROS~2~\cite{ros-robotics-companies}, combined with the computational improvements offered by hardware acceleration, accentuates the community's demand for a standardized, industry-grade benchmark to evaluate varied hardware solutions. Recently, there has been a plethora of workshops and tutorials focusing on benchmarking robotics applications~\cite{icra2021,icra2022-metrics,icra2022-sbd,iros2020,iros2021,iros2022,iros2023,rss2020,rss2021,rss2022,rss2023}, and while benchmarks for specific robotics algorithms~\cite{bakhshalipour2022rtrbench,neuman2019benchmarking} and certain end-to-end robotic applications, such as drones~\cite{boroujerdian2018mavbench,krishnan2022automatic,krishnan2022roofline,nikiforov2023rose}, do exist, the nuances of analyzing general ROS 2 computational graphs on heterogeneous hardware is yet to be fully understood.

\begin{figure}[tbp]
\vspace{.5em}
\centering
\includegraphics[scale=0.60]{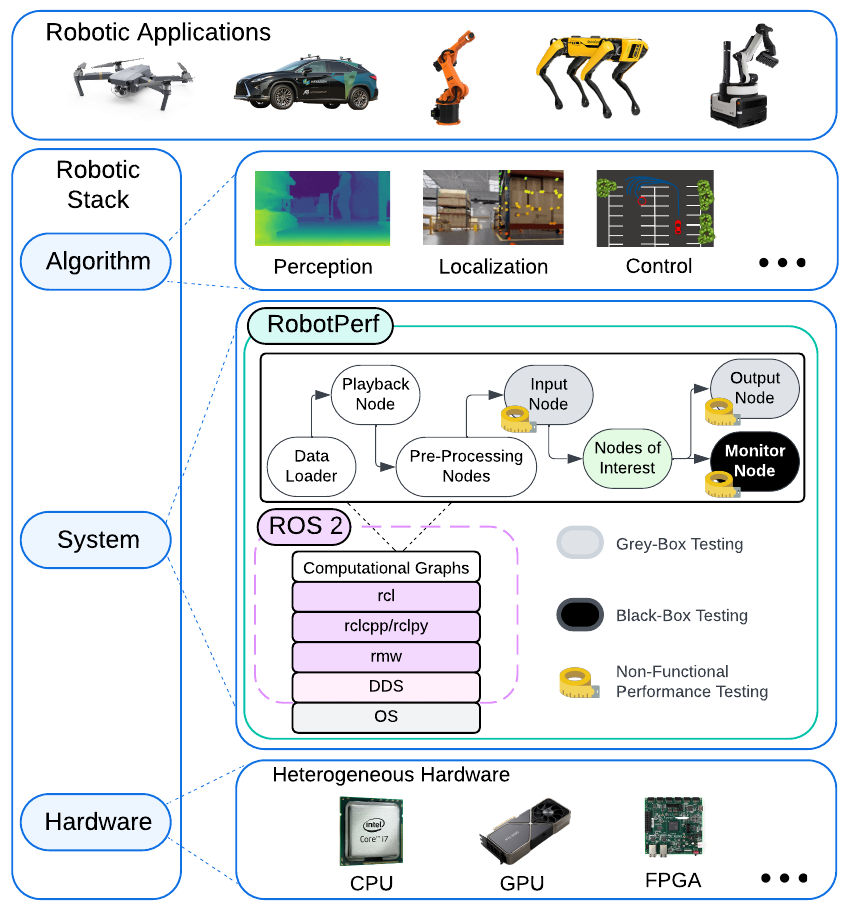}
\caption{A high level overview of RobotPerf. It targets industry-grade real-time systems with complex and extensible computation graphs using the Robot Operating System (ROS~2) as its common baseline. Emphasizing adaptability, portability, and a community-driven approach, RobotPerf aims to provide fair comparisons of ROS 2 computational graphs across CPUs, GPUs, FPGAs and other accelerators.}
\vspace{-18.5pt}
\label{fig:robotperf_overview}
\end{figure}



In this paper, we introduce \textit{RobotPerf}, an open-source and community-driven benchmarking tool designed to assess the performance of robotic computing systems in a standardized, architecture-neutral, and reproducible way, accommodating the various combinations of hardware and software in different robotic platforms (see Figure~\ref{fig:robotperf_overview}). RobotPerf focuses on evaluating robotic workloads in the form of ROS~2 computational graphs on a wide array of hardware setups, encompassing a complete robotics pipeline and emphasizing real-time critical metrics. The framework incorporates two distinct benchmarking methodologies that utilize various forms of instrumentation and ROS \emph{nodes} to capture critical metrics in robotic systems. These approaches are: black-box testing, which measures performance by eliminating upper layers and replacing them with a test application, and grey-box testing, an application-specific measure that observes internal system states with minimal interference. The framework is user-friendly, easily extendable for evaluating custom ROS 2 computational graphs, and collaborates with major hardware acceleration vendors for a standardized benchmarking approach. It aims to foster research and innovation as an open-source project. We validate the framework's capabilities by conducting benchmarks on diverse hardware platforms, including CPUs, GPUs, and FPGAs, thereby showcasing RobotPerf's utility in drawing valuable performance insights. 

RobotPerf's source code and documentation are available at \url{https://github.com/robotperf/benchmarks} and its methodologies are currently being used in industry to benchmark industry-strength, production-grade systems. 


%% file: tex/2_background.tex
\subsection{The Robot Operating System (ROS and ROS~2)}
ROS~\cite{quigley2009ros} is a widely-used middleware for robot development that serves as a \emph{structured communications layer} and offers a comprehensive suite of additional functionalities including: open-source packages and drivers for various tasks, sensors, and actuators, as well as a collection of tools that simplify development, deployment, and debugging processes. ROS enables the creation of computational graphs (see Figure \ref{fig:robotperf_overview}) that connect software processes, known as nodes, through topics, facilitating the development of end-to-end robotic systems. Within this framework, nodes can publish to or subscribe from topics, enhancing the modularity of robotic systems. 


%




ROS~2 builds upon ROS and addresses many of its key limitations. Constructed to be industry-grade, ROS~2 adheres to industry Data Distribution Service (DDS) and Real-Time Publish Subscribe (RTPS) standards~\cite{OMG-DDSI-RTPS-2.5}.
Based on the Data Distribution Service (DDS) standard, it enables fine-grained, direct, inter- and intra-node communication, enhancing performance, reducing latency, and improving scalability. Importantly, these improvements are also designed to support hardware acceleration~\cite{mayoral2021adaptivecomputing,mayoral2022robotcore}. Over 600 companies have adopted ROS~2 and its predecessor ROS in their production environments, underscoring its significance and widespread adoption in the industry \cite{ros-robotics-companies}.

ROS~2 also provides standardized APIs to connect user code through
language-specific client libraries, \textit{rclcpp} and \textit{rclpy}, which 
handle the scheduling and invocation of callbacks such as timers, subscriptions, and services. Without a ROS Master, ROS 2 creates a decentralized framework where nodes discover each other 
and manage their own parameters. 

%% file: tex/3_related.tex
\begin{table}
\vspace{7pt}
\centering
\begin{tabular}{lccccccccc}
\toprule
& \multicolumn{7}{c}{\textbf{Characteristics}} \\
\cmidrule(lr){2-8}
& \rotatebox{90}{\scriptsize Real-time Performance Metrics}
& \rotatebox{90}{\scriptsize Spans Multiple Pipeline Categories }
& \rotatebox{90}{\scriptsize Evaluation on Heterogeneous Hardware}
& \rotatebox{90}{\scriptsize Integration with ROS/ROS~2 Framework}
& \rotatebox{90}{\scriptsize Functional Performance Testing}
& \rotatebox{90}{\scriptsize Non-functional Performance Testing}
& \rotatebox{90}{\scriptsize Community Led} \\
\midrule
OMPL Benchmark \cite{sucan2012open} & \cmark  & \xmark & \xmark & \xmark & \xmark & \cmark & \xmark \\
MotionBenchMaker \cite{chamzas2021motionbenchmaker} & \cmark  & \xmark & \xmark & \xmark & \cmark & \cmark & \xmark \\
OpenCollBench \cite{tan2020opencollbench} & \xmark & \xmark & \cmark & \xmark & \cmark & \xmark & \xmark \\
BARN \cite{perille2020benchmarking} & \xmark  & \xmark & \xmark & \cmark & \cmark & \xmark & \xmark \\
DynaBARN \cite{nair7dynabarn} & \cmark & \xmark & \xmark & \cmark & \cmark & \xmark & \xmark \\
MAVBench \cite{boroujerdian2018mavbench} & \cmark  & \cmark & \cmark & \cmark & \cmark & \cmark & \xmark \\
Bench-MR \cite{heiden2021bench} & \cmark  & \xmark & \xmark & \xmark & \cmark & \xmark & \xmark \\
RTRBench \cite{bakhshalipour2022rtrbench} & \cmark  & \cmark & \xmark & \xmark & \xmark & \cmark & \xmark \\
\midrule
\textbf{RobotPerf} (\textbf{ours}) & \cmark  & \cmark & \cmark & \cmark & \xmark & \cmark & \cmark \\
\bottomrule
\end{tabular}
\caption{Comparative evaluation of representative existing robotics benchmarks with RobotPerf across essential characteristics for robotic systems. 
}
\label{tab:works}
\vspace{-20pt}
\end{table}

%
%
%
%
%
%
%
%

\subsection{Robotics Benchmarks}


There has been much recent development of open-source robotics libraries and associated benchmarks demonstrating their performance as well as a plethora of workshops and tutorials focusing on benchmarking robotics applications~\cite{icra2021,icra2022-metrics,icra2022-sbd,iros2020,iros2021,iros2022,iros2023,rss2020,rss2021,rss2022,rss2023}. However, most of these robotics benchmarks focus on algorithm correctness (\emph{functional} testing) in the context of domain specific problems, as well as end-to-end latency on CPUs~\cite{sucan2012open,chamzas2021motionbenchmaker,tan2020opencollbench,perille2020benchmarking,mollbenchmarking,nair7dynabarn,kingston2022robowflex,heiden2021bench,ahn2020robel,weisz2016robobench,del2006benchmarks,michel2008rat,murali2019pyrobot,james2020rlbench,leitner2017acrv,zhu2020robosuite,fan2018surreal,althoff2017commonroad}.
A few works also analyze some \emph{non-functional} metrics, such as CPU performance benchmarks, to explore bottleneck behaviors in selected workloads~\cite{bakhshalipour2022rtrbench,neuman2019benchmarking,delmerico2018benchmark}.


Recent work has also explored the implications of operating systems and task schedulers on ROS~2 computational graph performance through benchmarking~\cite{reke2020self,barut2021benchmarking,puck2020distributed,yang2020exploring,arafat2022response} as well as by optimizing the scheduling and communication layers of ROS and ROS~2 themselves~\cite{sugata2017acceleration,ohkawa2019high,choi2021picas,suzuki2018real,gutierrez2018time,gutierrez2018real,gutierrez2018towards,gutierrez2018synch}. These works often focused on a specific context or (set of) performance counter(s).

Finally, previous work has leveraged hardware acceleration for select ROS Nodes and adaptive computing to optimize the ROS computational graphs~\cite{yamashina2015proposal,yamashina2016crecomp,podlubne2019fpga,eisoldt2021reconfros,lienen2020reconros,9415584,ohkawa2016architecture,panadda2021low,8956928,9397897,ohkawa2018fpga,9355892,amano2021dataset,nitta2018study,chen2021fogros,nvidia2022isaacros,wan2022analyzing}. However, these works do not provide comprehensive frameworks to quickly analyze and evaluate new heterogeneous computational graphs except for two works that are limited to the context of UAVs~\cite{boroujerdian2018mavbench,nikiforov2023rose}.

Research efforts most closely related to our work include \texttt{ros2\_tracing} \cite{bedard2022ros2_tracing} and RobotCore \cite{mayoral2022robotcore}. \texttt{ros2\_tracing} provided instrumentation that demonstrated integration with the low-overhead LTTng tracer into ROS 2, while RobotCore illuminates the advantages of using vendor-specific tracing to complement \texttt{ros2\_tracing} to assess the performance of hardware-accelerated ROS 2 Nodes. 
Building on these two specific foundational contributions, RobotPerf offers a comprehensive set of ROS~2 kernels spanning the robotics pipeline and evaluates them on diverse hardware.

Table~\ref{tab:works} summarizes our unique contributions. It includes a selection of representative benchmarks from above and provides an evaluation of these benchmarks against RobotPerf, focusing on essential characteristics vital for robotic systems. We note that while our current approach focuses only on non-functional performance benchmarking tests, RobotPerf's architecture and methodology can be extended to also measure functional metrics.

%% file: tex/4_design.tex
\section{RobotPerf: Principles \& Methodology}

RobotPerf is an open-source, industry-strength robotics benchmark for portability across heterogeneous hardware platforms. This section outlines the important design principles and describes the implementation methodology.

\subsection{Non-Functional Performance Testing}


Currently, RobotPerf specializes in non-functional performance testing, evaluating the efficiency and operational characteristics of robotic systems. Non-functional performance testing measures those aspects not belonging to the system's functions, such as computational latency, memory consumption, and CPU usage. 
In contrast, traditional functional performance testing looks into the system's specific tasks and function, verifying its effectiveness in its primary goals, like the accuracy of the control algorithm in following a planned robot's path. While functional testing confirms a system performs its designated tasks correctly, non-functional testing ensures it operates efficiently and reliably.

\subsection{ROS 2 Integration \& Adaptability} 


RobotPerf is designed specifically to evaluate ROS~2 computational graphs, rather than focusing on independent robotic algorithms. We emphasize benchmarking \textit{ROS~2 workloads} because the use of ROS~2 as middleware allows for the easy composition of complex robotic systems. 
This makes the benchmark versatile and well-suited for a wide range of robotic applications and enables industry, which is widely using ROS, to rapidly adopt RobotPerf. 



\subsection{Platform Independence \& Portability} 
RobotPerf allows for the evaluation of benchmarks on a variety of hardware platforms, including general-purpose CPUs and GPUs, reconfigurable FPGAs, and specialized accelerators (e.g., ray tracing accelerators \cite{deng2017toward}). 
Benchmarking robotic workloads on heterogeneous platforms is vital to evaluate their respective capabilities and limitations. This facilitates optimizations for efficiency, speed, and adaptability, as well as fine-tuning of resource allocations, ensuring robust and responsive operation across diverse contexts.

\begin{table}[!t]
\vspace{8pt}
\centering
\footnotesize
\renewcommand{\arraystretch}{1.1} 
\setlength{\tabcolsep}{6pt} 
\begin{tabularx}{\columnwidth}{lXX}
\toprule
\textsc{\scriptsize Criteria} & \textbf{Grey-Box} & \textbf{Black-Box} \\
\midrule
\textsc{\scriptsize Precision} & Utilizes tracers from in-code instrumentation. & Limited to ROS 2 message subscriptions. \\
\textsc{\scriptsize Performance} & Low overhead. Driven by kernelspace. & Restricted to ROS 2 message callbacks. Recorded by userspace processes.  \\
\textsc{\scriptsize Flexibility} & Multiple event types. & Limited to message subscriptions in current implementation. \\
\textsc{\scriptsize Portability} & Requires a valid tracer. Standard format (CTF). & Standard ROS 2 APIs. Custom JSON format. \\
\textsc{\scriptsize Ease of use} & Requires code modifications and data postprocessing.
& Tests unmodified software with minor node additions. \\
\textsc{\scriptsize Real-Robots} & Does not modify the computational graph. & Modifies the computational graph adding extra dataflow. \\
\bottomrule
\end{tabularx}
\caption{Grey-box vs. black-box benchmarking trade-offs.}
\label{tab:tradeoffs}
\vspace{-18.5pt}
\end{table}

\subsection{Flexible Methodology}


We offer grey-box and black-box testing methods to suit different needs. Black-box testing provides a quick-to-enable external perspective and measures performance by eliminating the layers above the layer-of-interest and replacing those with a specific test application. Grey-box testing provides more granularity and dives deeper into the internal workings of ROS 2, allowing users to generate more accurate measurements at the cost of increased engineering effort. 
As such, each method has its trade-offs, and providing both options enables users flexibility. We describe each method in more detail below and highlight takeaways in Table~\ref{tab:tradeoffs}.

\subsubsection{Grey-Box Testing}

Grey-box testing enables precise probe placement within a robot's computational graph, 
generating a chronologically ordered log of critical events using a tracer that could be proprietary or open source, such as LTTng~\cite{desnoyers2006lttng}. As this approach is fully integrated with standard ROS 2 layers and tools through \texttt{ros2\_tracing}, it incurs a minimal average latency of only 3.3 \si{\micro\second}~\cite{bedard2022ros2_tracing}, making it well-suited for real-time systems. 
With this approach, optionally, RobotPerf offers specialized input and output nodes that are positioned outside the nodes of interest to avoid the need to 
instrument them.
These nodes generate the message tracepoints upon publish and subscribe events 
which are processed to calculate end-to-end latency. 

\subsubsection{Black-Box Testing}

The black-box methodology utilizes a user-level node called the \texttt{MonitorNode} to evaluate the performance of a ROS~2 node. 
The \texttt{MonitorNode} subscribes to the target node, recording the timestamp when each message is received. By accessing the propagated ID, the \texttt{MonitorNode} determines the end-to-end latency by comparing its timestamp against the \texttt{PlaybackNode}'s recorded timestamp for each message. While this approach does not need extra instrumentation, and is easier to implement, it offers a less detailed analysis and alters the computational graph by introducing new nodes and dataflow.

\input{tex/benchmark_table}

\subsection{Opaque Performance Tests} 

The requirement for packages to be instrumented directly within the source code poses a challenge to many benchmarking efforts. To overcome this hurdle, for most benchmarks, we refrain from altering the workloads of interest and, instead, utilize specialized input and output nodes positioned outside the primary nodes of concern. This setup allows for benchmarking without the need for direct instrumentation of the target layer. We term this methodology ``opaque tests," a concept that RobotPerf adheres to when possible. 

\subsection{Reproducibility \& Consistency} 
To ensure consistent and reproducible evaluations, RobotPerf adheres to specific common robotic dataformats. In particular, it uses ROS 2 \texttt{rosbags}, including our own available at \url{https://github.com/robotperf/rosbags}, as well third-party bags (e.g., the \texttt{r2b} dataset~\cite{nvidia_r2bdataset2023}). 


To ensure consistent data loading and finer control over message delivery rates, we drew inspiration from~\cite{ros2_benchmark}. Our computational graphs incorporate \emph{modified and improved} \texttt{DataLoaderNode} and \texttt{PlaybackNode} implementations, which can be accessed at \url{https://github.com/robotperf/ros2_benchmark}. These enhanced nodes offer improvements that report  worst-case latency and 
enable the reporting of maximum latency, introduce the ability to profile power consumption and so forth.

\subsection{Metrics}


We focus on three key metrics: latency, throughput and power consumption including energy efficiency. 
Latency measures the time between the start and the completion of a task. Throughput measures the total amount of work done in a given time for a task. Power measures the electrical energy per unit of time consumed while executing a given task. Measuring energy efficiency (or performance-per-Watt) captures the total amount of work (relative to either throughput or latency) that can be delivered for every watt of power consumed and is directly related to the runtime of battery powered robots~\cite{boroujerdian2018mavbench}. 

\subsection{Current Benchmarks and Categories}

RobotPerf \texttt{beta}~\cite{robotperf_benchmarks_repo} introduces benchmarks that cover the robotics pipeline from perception, to localization, to control, as well as dedicated benchmarks for manipulation. 
The full list of benchmarks in the \texttt{beta} release can be found in Table~\ref{tab:benchmarks_table}. 
Aligned with our principles defined above, each benchmark is a self-contained ROS 2 package which describes all dependencies (generally other ROS packages). To facilitate reproducibility, all benchmarks are designed to be built and run using the common ROS 2 development flows (\texttt{ament} build tools, \texttt{colcon} meta-build tools, etc.).
Finally, so that the benchmarks can be easily consumed by other tools, a description of each benchmark, as well as its results, is defined in a machine-readable format. 
As such, accompanying the \texttt{package.xml} and \texttt{CMakeLists.txt} files required for all ROS packages, a YAML file named \texttt{benchmark.yaml} is  in the root of each benchmark which  describes the benchmark and includes  accepted results.

\subsection{Run Rules}

To ensure the reliability and reproducibility of the performance data, we adhere to a stringent set of run rules. 
First, tests are performed in a controlled environment to ensure that performance data is not compromised by fluctuating external parameters. 
As per best practices recommended by \texttt{ros2\_tracing}~\cite{bedard2022ros2_tracing}, we record and report settings like clock frequency and core count.
%
%
Second, we look forward to the possibility of RobotPerf being embraced by the community and have results undergo peer review, which can contribute to enhancing reproducibility and accuracy. 
%
%
Finally, we aim to avoid overfitting to specific hardware setups or software configurations by encompassing a broad spectrum of test scenarios. 

%% file: tex/benchmark_table.tex
\begin{table*}[!t]
\vspace{7pt}
\centering
\renewcommand{\arraystretch}{1.2} 
\begin{tabularx}{\textwidth}{lllX}
\toprule
\textbf{Category} & \textbf{Benchmark Name} & \textbf{Description} \\
\midrule
\multirow{8}{*}{\textbf{Perception}} 
& \href{https://github.com/robotperf/benchmarks/tree/main/benchmarks/perception/a1_perception_2nodes}{a1\_perception\_2nodes} & Graph with 2 components: \texttt{rectify} and \texttt{resize} \cite{ros_acceleration_2023, ros_perception_image_proc_2023}. \\
& \href{https://github.com/robotperf/benchmarks/tree/main/benchmarks/perception/a2_rectify}{a2\_rectify} & \texttt{rectify} component \cite{ros_acceleration_2023, ros_perception_image_proc_2023}. \\
& \href{https://github.com/robotperf/benchmarks/tree/main/benchmarks/perception/a3_stereo_image_proc}{a3\_stereo\_image\_proc} & Computes disparity map from left and right images \cite{ros_perception_stereo_image_proc_2023}. \\
& \href{https://github.com/robotperf/benchmarks/tree/main/benchmarks/perception/a4_depth_image_proc}{a4\_depth\_image\_proc} & Computes point cloud from rectified depth and color images \cite{ros_perception_depth_image_proc_2023}. \\
& \href{https://github.com/robotperf/benchmarks/tree/main/benchmarks/perception/a5_resize}{a5\_resize} & \texttt{resize} component \cite{ros_acceleration_2023, ros_perception_image_proc_2023}. \\
\midrule
\multirow{3}{*}{\textbf{Localization}} 
& \href{https://github.com/robotperf/benchmarks/tree/main/benchmarks/localization/b1_visual_slam}{b1\_visual\_slam} & Visual SLAM component \cite{nvidia_isaac_ros_2023_vslam}. \\
& \href{https://github.com/robotperf/benchmarks/tree/main/benchmarks/localization/b2_map_localization}{b2\_map\_localization} & Map localization component \cite{nvidia_isaac_ros_map_localization_2023}. \\
& \href{https://github.com/robotperf/benchmarks/tree/main/benchmarks/localization/b3_apriltag_detection}{b3\_apriltag\_detection} & Apriltag detection component \cite{nvidia_isaac_ros_apriltag_2023}. \\
\midrule
\multirow{5}{*}{\textbf{Control}} 
& \href{https://github.com/robotperf/benchmarks/tree/main/benchmarks/control/c1_rrbot_joint_trajectory_controller}{c1\_rrbot\_joint\_trajectory\_controller} & Joint trajectory controller \cite{ros2_controllers_joint_2023}. \\
& \href{https://github.com/robotperf/benchmarks/tree/main/benchmarks/control/c2_diffbot_diff_driver_controller}{c2\_diffbot\_diff\_driver\_controller} & Differential driver controller \cite{ros2_controllers_diff_drive_2023}. \\
& \href{https://github.com/robotperf/benchmarks/tree/main/benchmarks/control/c3_rrbot_forward_command_controller_position}{c3\_rrbot\_forward\_command\_controller\_position} & Position-based forward command controller \cite{ros2_controllers_forward_command_2023}. \\
& \href{https://github.com/robotperf/benchmarks/tree/main/benchmarks/control/c4_rrbot_forward_command_controller_velocity}{c4\_rrbot\_forward\_command\_controller\_velocity} & Velocity-based forward command controller \cite{ros2_controllers_forward_command_2023}. \\
& \href{https://github.com/robotperf/benchmarks/tree/main/benchmarks/control/c5_rrbot_forward_command_controller_acceleration}{c5\_rrbot\_forward\_command\_controller\_acceleration}  & Acceleration-based forward command controller \cite{ros2_controllers_forward_command_2023}. \\
\midrule
\multirow{6}{*}{\textbf{Manipulation}} 
& \href{https://github.com/robotperf/benchmarks/tree/main/benchmarks/manipulation/d1_xarm6_planning_and_traj_execution}{d1\_xarm6\_planning\_and\_traj\_execution} & Manipulator planning and trajectory execution \cite{moveit_ros_2023}. \\
& \href{https://github.com/robotperf/benchmarks/tree/main/benchmarks/manipulation/d2_collision_checking_fcl}{d2\_collision\_checking\_fcl} & Collision check: manipulator and box (FCL \cite{fcl2023}). \\
& \href{https://github.com/robotperf/benchmarks/tree/main/benchmarks/manipulation/d3_collision_checking_bullet}{d3\_collision\_checking\_bullet}  & Collision check: manipulator and box (Bullet \cite{coumans2021}). \\
& \href{https://github.com/robotperf/benchmarks/tree/main/benchmarks/manipulation/d4_inverse_kinematics_kdl}{d4\_inverse\_kinematics\_kdl}  & Inverse kinematics (KDL plugin \cite{moveit_kdl_2023}). \\
& \href{https://github.com/robotperf/benchmarks/tree/main/benchmarks/manipulation/d5_inverse_kinematics_lma}{d5\_inverse\_kinematics\_lma} & Inverse kinematics (LMA plugin \cite{moveit2023_lma}). \\
& \href{https://github.com/robotperf/benchmarks/tree/main/benchmarks/manipulation/d6_direct_kinematics}{d6\_direct\_kinematics} & Direct kinematics for manipulator \cite{moveit_ros_2023}. \\
\bottomrule
\end{tabularx}
\caption{RobotPerf \texttt{beta} Benchmarks (see \cite{robotperf_benchmarks_repo}).}
\label{tab:benchmarks_table}
\vspace{-20pt}
\end{table*}

%% file: tex/5_eval.tex

%


We conduct comprehensive benchmarking using RobotPerf to evaluate its capabilities on three key aspects vital for a robotics-focused computing benchmark. First, we validate the framework's capacity to provide comparative insights across divergent heterogeneous platforms from edge devices to server-class hardware. Second, we analyze the results to understand RobotPerf's ability to guide selection of the optimal hardware solution tailored to particular robotic workloads. Finally, we assess how effectively RobotPerf reveals the advantages conferred by hardware and software acceleration techniques relative to general-purpose alternatives. All of our results and source code can be found open-source at: \url{https://github.com/robotperf/benchmarks}.

\subsection{Fair and Representative Assessment of Heterogeneity}

Assessing hardware heterogeneity in robotic applications is imperative in the ever-evolving field of robotics. Different robotic workloads demand varying computational resources and efficiency levels. Therefore, comprehensively evaluating performance across diverse hardware platforms
is crucial. 

We evaluated the RobotPerf benchmarks over a wide list of hardware platforms, including general-purpose CPUs on edge devices (e.g., Qualcomm RB5), server-class CPUs (e.g., Intel i7-8700), and specialized hardware accelerators (e.g., AMD Kria KR260). Figure \ref{fig:colors_hardware} illustrates benchmark performance in robotics per category of workload (perception, localization, control, and manipulation) using radar plots, wherein the different hardware solutions are depicted together alongside different robotic workloads per category.  Each hardware solution is presented with a different color, with smaller values and areas representing better performance in the respective category. 
%
%
Given our ability to benchmark 18 platforms (bottom of Figure~\ref{fig:colors_hardware}), RobotPerf is capable of benchmarking heterogeneous hardware platforms and workloads, 
paving the way for community-driven co-design and optimization of hardware and software. 



\subsection{Quantitative Approach to Hardware Selection}



The rapid evolution and diversity of tasks in robotics means we need to have a meticulous and context-specific approach to computing hardware selection and optimization. A ``one-size-fits-all'' hardware strategy would be an easy default selection, but it fails to capitalize on the nuanced differences in workload demands across diverse facets like perception, localization, control, and manipulation, each exhibiting distinctive sensitivities to hardware capabilities. Therefore, a rigorous analysis, guided by tools like RobotPerf, becomes essential to pinpoint the most effective hardware configurations that align well with individual workload requirements. 

The results in Figure \ref{fig:colors_hardware} demonstrate the fallacy of a ``one-size-fits-all'' solution. 
For example, focusing in on the latency radar plot for control from Figure \ref{fig:colors_hardware} (col 3, row 1), we see that the i7-12700H (I7H) outperforms the NVIDIA AGX Orin Dev. Kit (NO) on benchmarks \texttt{C1}, \texttt{C3}, \texttt{C4}, and \texttt{C5}, but is $6.5\times$ slower on benchmark \texttt{C2}.
%
As such, by analyzing data from the RobotPerf benchmarks, roboticists can better determine which hardware option best suits their needs given their specific workloads and performance requirements.

One general lesson learned while evaluating the data is that each workload is unique, making it hard to generalize across both benchmarks and categories. To that end, RobotPerf results help us understand how the use of various hardware solutions and dedicated domain-specific hardware accelerators significantly improves the performance.

\begin{figure}[!t]
\centering
\includegraphics[width=.9\columnwidth]{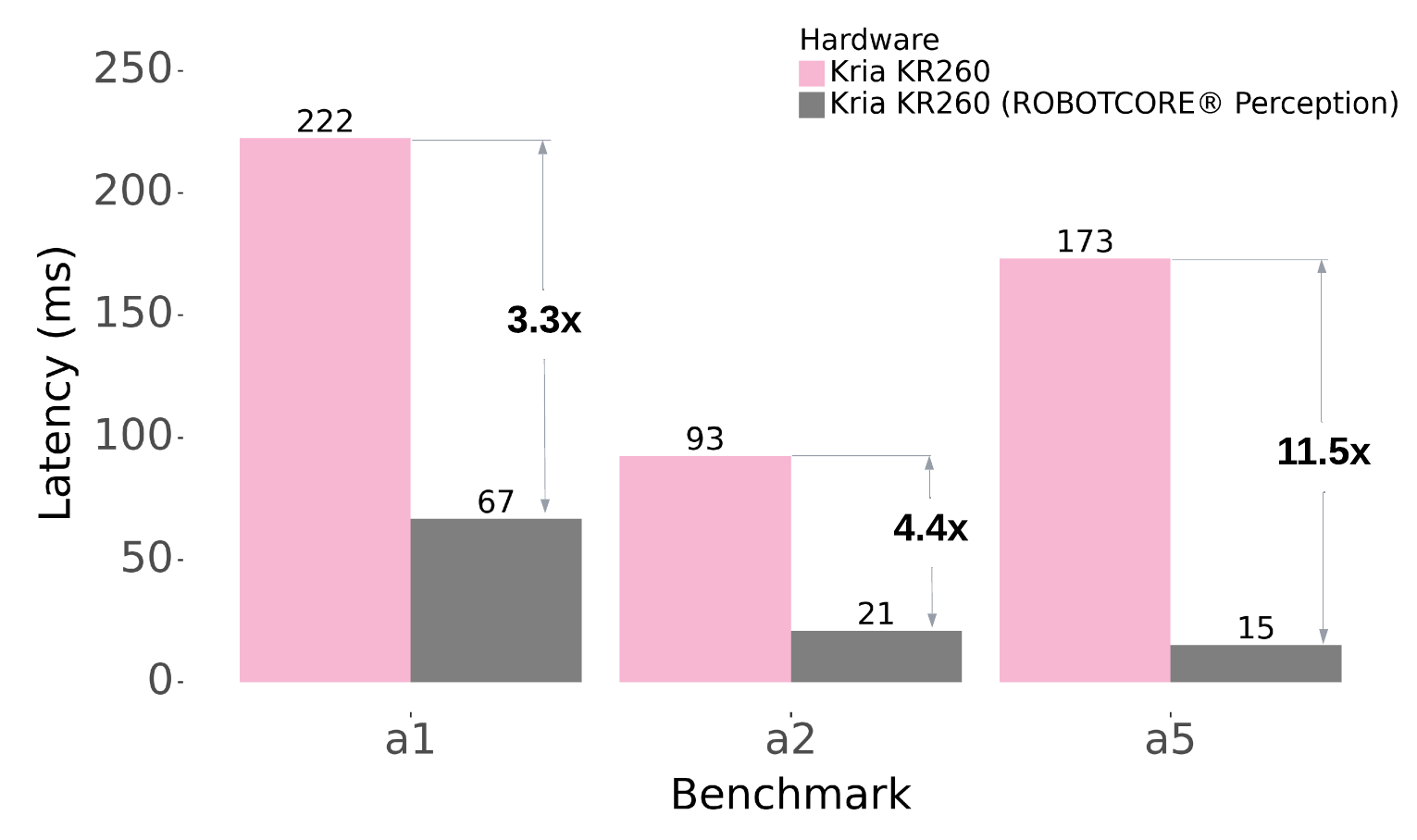}
\vspace{-5pt}
\caption{Benchmark comparison of perception latency (ms) on AMD's Kria KR260 with and without the ROBOTCORE Perception accelerator. The benchmarks used are \texttt{a1}, \texttt{a2}, and \texttt{a5} as defined in Table~\ref{tab:benchmarks_table}. We find that hardware acceleration can enable performance gains of as much as 11.5$\times$.}
\vspace{-1em}
\label{fig:robotperf_perception_comparison_KR260}
\end{figure}

\subsection{Rigorous Assessment of Acceleration Benefits}

In the rapidly advancing field of computing hardware, the optimization of algorithm implementations 
is a crucial factor in determining the success and efficiency of robotic applications. The need for an analytical tool, like RobotPerf, that facilitates the comparison of various algorithmic implementations on uniform hardware setups becomes important. 

Figure \ref{fig:robotperf_perception_comparison_KR260} is a simplified version of Figure \ref{fig:colors_hardware}, depicting AMD's Kria KR260 hardware solution in two forms: the usual hardware and a variant that leverages a domain-specific hardware accelerator (ROBOTCORE Perception, a soft-core running in the FPGA for accelerating perception robotic computations). 
The figure demonstrates that hardware acceleration can enable performance gains of as much as 11.5$\times$ (from 173~ms down to 15~ms for benchmark \texttt{a5}). 
We stress that the results obtained here should be interpreted according to each end application and do not represent a generic recommendation on which hardware should be used. Other factors, including availability, the form factor, and community support, are relevant aspects to consider when selecting a hardware solution.




\section{Conclusion and Future Work}\label{sec:conclusion}
\input{tex/6_conclusion}

\input{tex/full_page_figure}

%% file: tex/6_conclusion.tex
RobotPerf represents an important step towards standardized benchmarking in robotics. With its comprehensive evaluation across the hardware/software stack and focus on industry-grade ROS 2 deployments, RobotPerf can pave the way for rigorous co-design of robotic hardware and algorithms. As RobotPerf matures with community involvement, we expect it to compare CPU, GPU and FPGA, exploring their power consumption and flexibility in augmenting real-world robotic computations. With a standardized robotics benchmark as a focal point, the field can make rapid progress in delivering real-time capable systems that will unlock the true potential of robotics in real-world applications. 

%% file: tex/full_page_figure.tex
\begin{figure*}[!t]
\centering
\vspace{2em}
\includegraphics[width=\textwidth]{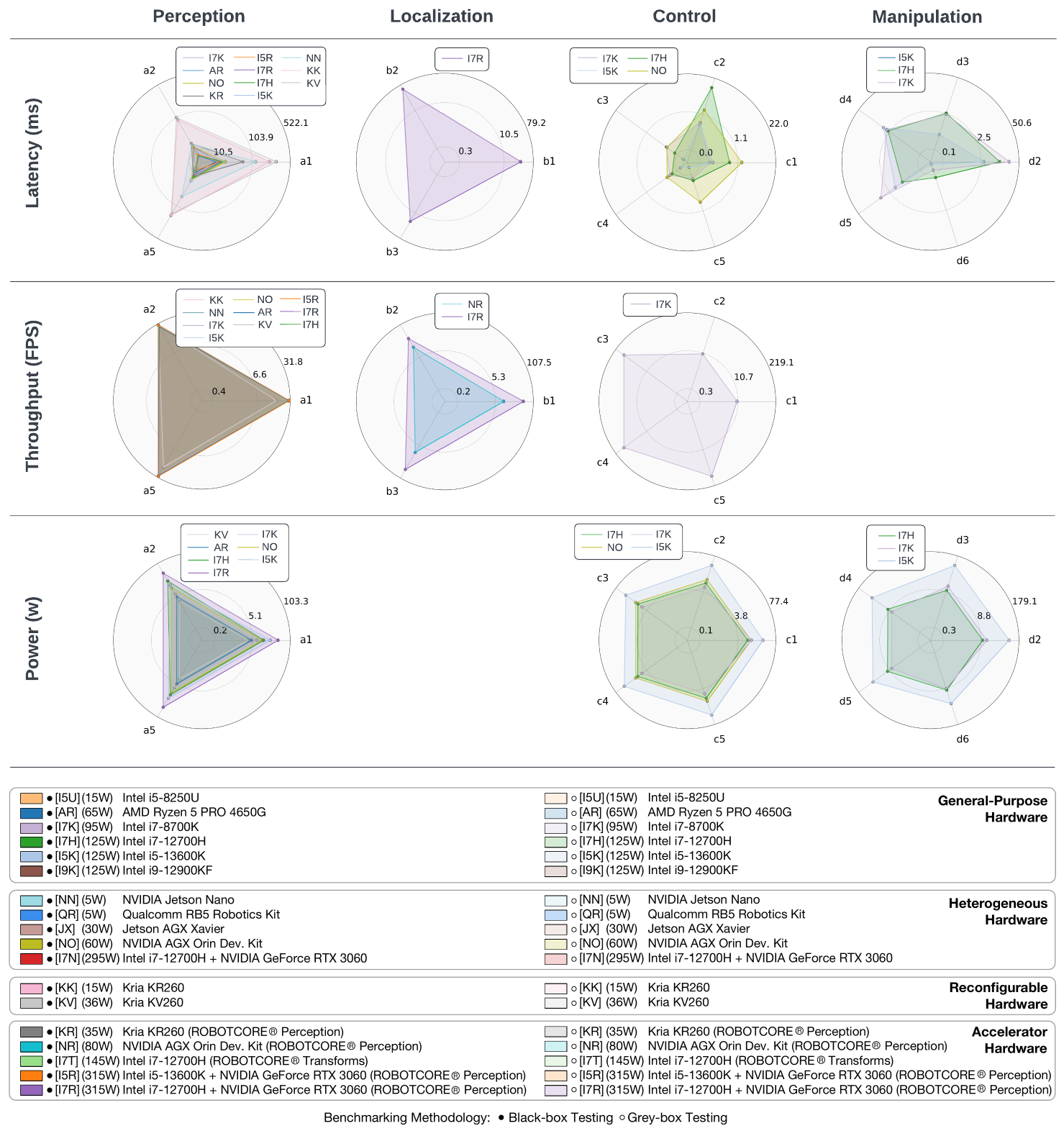}  

\caption{Benchmarking results on diverse hardware platforms across perception, localization, control, and manipulation workloads defined in RobotPerf \texttt{beta} Benchmarks. Radar plots illustrate the latency, throughput, and power consumption for each hardware solution and workload, with reported values representing the maximum across a series of runs. The labels of vertices represent the workloads defined in Table \ref{tab:benchmarks_table}. Each hardware platform and performance testing procedure is delineated by a separate color, with darker colors representing Black-box testing and lighter colors Grey-box testing. In the figure's key, the hardware platforms are categorized into four specific types: general-purpose hardware, heterogeneous hardware, reconfigurable hardware, and accelerator hardware. Within each category, the platforms are ranked based on their Thermal Design Power (TDP), which indicates the maximum power they can draw under load. The throughput values for manipulation tasks and power values for localization tasks have not been incorporated into the \texttt{beta} version of RobotPerf. As RobotPerf continues to evolve, more results will be added in subsequent iterations.
}
\label{fig:colors_hardware}
\end{figure*}